# One-shot Localization and Segmentation of Medical Images with Foundation Models

Deepa Anand    Gurunath Reddy M    Vanika Singhal    Dattesh D. Shanbhag

Shriram KS    Uday Patil    Chitresh Bhushan    Kavitha Manickam    Dawei Gui

Rakesh Mullick    Avinash Gopal    Parminder Bhatia    Taha Kass-Hout

GE HealthCare deepa.anand1@ge.com

## Abstract

Recent advances in Vision Transformers (ViT) and Stable Diffusion (SD) models with their ability to capture rich semantic features of the image have been used for image correspondence tasks on natural images. In this paper, we examine the ability of a variety of pre-trained ViT (DINO, DINOv2, SAM, CLIP) and SD models, trained exclusively on natural images, for solving the correspondence problems on medical images. While many works have made a case for in-domain training, we show that the models trained on natural images can offer good performance on medical images across different modalities (CT,MR,Ultrasound) sourced from various manufacturers, over multiple anatomical regions (brain, thorax, abdomen, extremities), and on wide variety of tasks. Further, we leverage the correspondence with respect to a template image to prompt a Segment Anything (SAM) model to arrive at single shot segmentation, achieving dice range of 62%-90% across tasks, using just one image as reference. We also show that our single-shot method outperforms the recently proposed few-shot segmentation method - UniverSeg (Dice range 47%-80%) on most of the semantic segmentation tasks(six out of seven) across medical imaging modalities.

## 1  Introduction

Foundation models, both self-supervised (DINO [3] [9], Stable Diffusion [12] and supervised (SAM [7], CLIP [11] have advanced the state of the art in Computer Vision. Most of these models benefit from deriving knowledge from tens of millions of natural images and text data. But, there is a hesitation in using these models on medical images due to the difference in acquisition physics and reconstruction algorithms [8]. Consequently, these models undergo expensive fine-tuning on limited relevant medical imaging data [5]. This study presents a methodology for one shot localization and segmentation of regions of interest for medical images, leveraging the existing FMs and chaining them as necessary (e.g. DINO with SAM) to obtain robust performance, without the need to perform any kind of fine-tuning on medical imaging data. We also present a comprehensive evaluation of the efficacy of various FMs (Dino V1/V2, Stable Diffusion, SAM, CLIP) on different datasets spanning a variety of tasks across different anatomies from multiple modalities (CT, MR, U/S) with images sourced from different manufacturers, and analyze their strengths and weaknesses. In addition, we

Preprint. Under review.

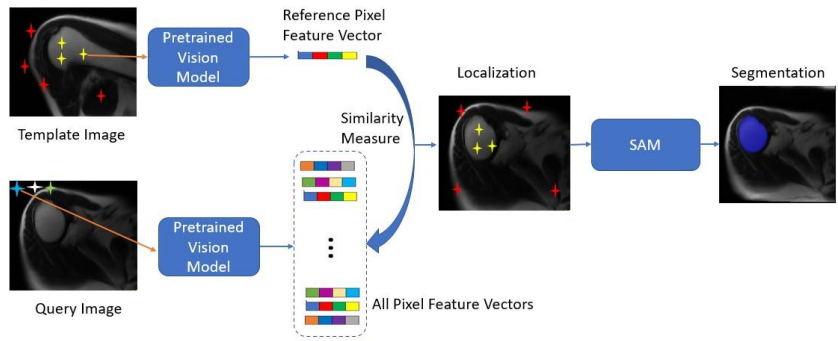

Figure 1: Proposed localization and segmentation framework

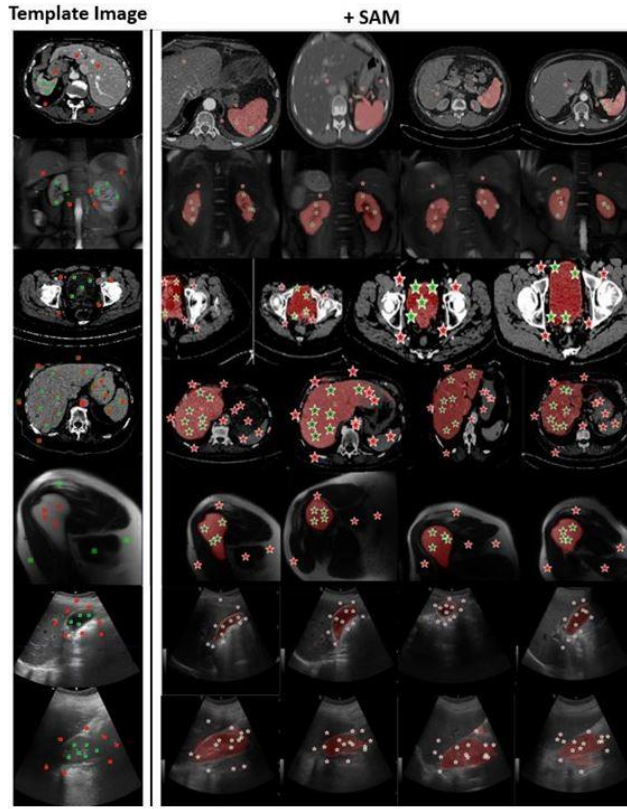

Figure 2: The results of chained approach (Feature based correspondence) followed by SAM based segmentation are shown for different anatomies and modalities. The left column shows the template images along with positive and negative prompts, while the right column shows predicted prompts along with the SAM segmentation masks for different target images.

compare our methods with the state-of-the-art medical few shot segmentation method UniverSeg [2] and demonstrate the superiority of the proposed approach on most of the segmentation tasks.

## 2 Literature Review

One of the earliest works using self-supervised learning (SSL) for automatic segmentation was presented in DINO v1. The authors showed that the attention masks from various heads had the ability to segregate semantic parts of an image but our experiments show a limited utility of this approach for medical images 6.2.1. Authors in [6] showed that models trained using DINO demonstrated emergence in their ability to extract patch level features for semantic segmentation. We deduce that such an approach can be very effective for region correspondence tasks with user interaction. Other works such as [1, 12, 14] demonstrate improvements in ability to obtain dense image features and their enhancements using FMs for correspondence tasks.

Segment Anything Model (SAM) [7] is a "prompt"-able (points, box and text) semantic segmentation model which we propose can be chained with the above mentioned dense feature based methods.UniverSeg is a new entrant in FM based semantic segmentation, offering the capability for few-shot segmentation using templates. We have compared our FM based correspondencesegmentation pipeline with UniverSeg (one shot and five shot).

### 2.1 Contributions

The main contributions of this paper are (a) chaining of outputs from exising FMs for medical image segmentation (b) ability for user to specify regions of interest(RoI) (c) comprehensive evaluation of different FMs for twelve different medical tasks

## 3 Methods

In the following sections we present our localization and chained segmentation framework and detail the methods to obtain dense features using the Foundation Models

### 3.1 Localization Framework

For the localization followed(optionally) by the segmentation framework, we assume the presence of a template image $T$ on which we mark the key positive points set (prompts) $T^+ = \{T_i^+, i = 1,...,n\}$ and negative point set $T^- = \{T_i^-, i = 1,..,m\}$. Positive point set denote the landmarks(for localization) or points in the region of interest. Let $I \in R^{P \times Q}$ be the target image on which localization or segmentation is to be performed.

$$Corr(T_i^+, I) = \underset{p,q}{\mathrm{argmax}}\, cosine\_similarity(T_i^{feat}, I_{p,q}^{feat}) \quad (1)$$

where, $p = 1,...P$ and $q = 1,...Q$. Broadly, we used the following methodology for one-shot localization: (a) First, the patch features are extracted and interpolated at pixel level from the template image denoted as $T_i^{feat}$, (b) correlation of features between template and target image is computed as given by equation 1. Where, target image features at pixel location $p,q$ are denoted by $I_{p,q}^{feat}$, (c) For landmark correspondence, this is the endpoint, (d) For organ segmentation, we utilize the landmark correspondence and any additional positive and negative prompts provided by user on template as a prompt for Segment Anything model (SAM). This entire framework is outlined in fig 1.

### 3.2 Pre-trained FM models for feature extraction

We utilized the following vision transformer models for dense feature extraction purposes: DINOv1 [3], DINOv2 [9], Segment Anything (SAM, ViT-H model) encoder [7] and Contrastive Language–Image Pre-training (CLIP, ViT-B model) vision encoder [11]. For DINO, we used multiple different model configurations as follows:(a)Overall size (SZ) of the model:small(s), base(b), large(l) and gigantic(g)(b)Patch size (PS) used for image tokenization: 8, 14 and 16 (c)To facilitate easy reading, DINO models are abbreviated as : d[VERSION][SZ][PS] (d)For DINO, have following configurations used in experiments : d1s8, d1s16, d1b8, d1b16, d2s14, d2b14, d2l14 and d2g14. For feature extraction from diffusion models, we follow the method prescribed in [12] to derive image features from diffusion model using the pretrained Stable Diffusion (SD) model v.2.1.

### 3.3 Patch descriptors using Foundation Models

Vision Transformers work by dividing an image into non-overlapping patches of a certain size. The self-attention mechanism and the projection of tokens into key, query, value and token, allows us to derive multiple embeddings for patches as well as for overall image (CLS token). Unlike [1], where the "key embedding" is used for dense feature extraction, we find merit in using the "token embedding" from the last transformer block as the patch feature descriptor. Specifically for d1s8, d1s16, d1b8, d1b16, d2s14, d2b14 models, we use 11th layer tokens as the embeddings. For d2l14 and d2g14, we use the 23rd and the 39th layer token embeddings respectively. For SAM, we derive the patch embeddings as the output of the encoder layer and for CLIP we utilize the patch embeddings from the visual transformer model.

| Dataset Name | # Slices | Modalities | Target Task |
|---|---|---|---|
| TS:Urinary Bladder | 40 | CT | Segment urinary bladder |
| TS:Spleen | 69 | CT | Segment spleen |
| TS:Liver | 91 | CT | Segment liver |
| Kidney | 50 | MR (T2w coronal images) | Segment Kidney |
| Shoulder | 25 | MR (Sagittal Localizer images) | Segment Shoulder |
| Knee Axial | 283 | MR (Axial Localizer images) | Localize landmarks |
| Knee Sagittal | 283 | MR (Sagittal Localizer images) | Localize Landmarks |
| Spine:Coccyx | 182 | MR (T1w sagittal images) | Vertebrae labeling |
| Head Neck | 42 | CT (Head neck) | Eyeball and optic nerve localization |
| Kidney (US) | 66 | Ultrasound (Kidney) | Kidney Segmentation |
| Gall Bladder (US) | 29 | Ultrasound (Gall Bladder) | Gall Bladder Segmentation |

Table 1: Dataset description

| NED | MR Coccyx | Knee Axial | Head Neck | Knee Sagittal |
|---|---|---|---|---|
| d1b16 | 0.098 | 0.027 | 0.014 | 0.064 |
| d1b8 | 0.114 | 0.024 | 0.008 | 0.060 |
| d1s16 | 0.118 | 0.025 | 0.014 | 0.063 |
| d1s8 | 0.122 | 0.023 | 0.008 | 0.056 |
| d2b14 | 0.072 | 0.024 | 0.011 | 0.069 |
| d2g14 | 0.052 | 0.024 | 0.010 | 0.062 |
| d2l14 | 0.085 | 0.024 | 0.011 | 0.073 |
| d2s14 | 0.107 | 0.024 | 0.012 | 0.064 |
| SD | 0.034 | 0.020 | 0.007 | 0.059 |
| SAM | 0.120 | 0.028 | 0.010 | 0.134 |

Table 2: Normalized Euclidean distance(NED) [3] for various models and landmarks. If multiple landmarks present in a anatomy (e.g. knee MRI, these are averaged and reported). A value > 0.1 is considered as worse performance, while value < 0.05 is considered acceptable for clinical usage.

To improve the resolution of patch-level features, we adopted the method from [1], where the authors modified the existing non-overlapping patch generation to create overlapping patches along with the positional encodings of ViT to improve the spatial resolution. These are then interpolated to get the pixel level features. In addition, the neighborhood log binning [1] is used to enrich each patch feature with the descriptors from its context.

For deriving dense features using diffusion models, we follow the method prescribed in [12]. Diffusion models are generative and progressively refine a noisy image into a high-quality one, simulating the diffusion of information in a stochastic manner. The model is trained through a forward step, to predict the noise in an image given the noisy image and time step t, which can be leveraged to generate new images by iteratively refining noisy images. For the purpose of feature extraction, the input to the network at time step t is constructed by introducing appropriate level of noise and extracting the intermediate layer features as diffusion features(DIFT) [12]. This step is repeated multiple times, to enhance the stability of the features, and results aggregated. We follow the default hyper-parameters settings as designated in code repository corresponding to [12].

## 4 EXPERIMENTS AND RESULTS

In this section we examine the efficacy of variants of DINO [3], Dinov2 [9], SAM [7] and Stable Diffusion [12] for various segmentation and localization tasks. We do not consider CLIP [11] for these experiments since our initial experiments 6.2.3 to visually assess quality of dense features was not as promising as the others. For segmentation tasks we also compare against the state-of-the-art for few-shot segmentation [2].

### 4.1 Data

We evaluate the proposed one-shot segmentation and localization pipeline on a variety of CT, MR and ultrasound images using both in-house and publicly available datasets for a variety of tasks. A summary of the datasets used along with the tasks and other details is provided in Table 1. Additional details about the datasets can be found in supplementary materials 6.1.

All evaluation was done using 2D slices for all the tasks. For landmark localization, a template image with sample landmark location was used, while for segmentation tasks, the template image for each of the tasks used positive and negative prompts inside and outside the region of interest (ROI) to generate the pixel level features using different pre-trained foundation models.

Ground-truth (GT) marking For opensource datasets, GT provided by organizers was used as is. For in-house datasets, GT was marked by a set of radiologists/ technicians/ clinicians and verified by a Senior Radiologist using a dashboard.

Table 3: Prompt accuracy 2 for a given ROI segmentation. An accuracy of 1 is desirable. The accuracy is correlated to outcome Dice score.

| Accuracy | Kidney (MR) | | TS: Liver | | Shoulder | | TS: Urinary Bladder | | TS: Spleen | | U S GB | | U S Kidney | |
|---|---|---|---|---|---|---|---|---|---|---|---|---|---|---|
| | Pos | Neg | Pos | Neg | Pos | Neg | Pos | Neg | Pos | Neg | Pos | Neg | Pos | Neg |
| d1b16 | 0.863 | 1 | 0.972 | 0.916 | 0.840 | 1 | 0.725 | 1 | 0.652 | 0.818 | 0.577 | 0.9 | 0.696 | 0.863 |
| d1b8 | 0.883 | 0.990 | 0.981 | 0.818 | 0.860 | 0.980 | 0.817 | 0.993 | 0.826 | 0.822 | 0.922 | 0.969 | 0.619 | 0.912 |
| d1s16 | 0.816 | 0.990 | 0.951 | 0.913 | 0.830 | 1 | 0.676 | 1 | 0.690 | 0.822 | 0.534 | 0.922 | 0.692 | 0.85 |
| d1s8 | 0.860 | 0.960 | 0.978 | 0.858 | 0.930 | 0.990 | 0.786 | 0.993 | 0.792 | 0.851 | 0.905 | 0.961 | 0.66 | 0.882 |
| d2b14 | 0.920 | 0.930 | 0.940 | 0.967 | 0.820 | 0.990 | 0.658 | 0.993 | 0.661 | 0.840 | 0.913 | 0.956 | 0.69 | 0.918 |
| d2g14 | 0.933 | 1 | 0.945 | 0.948 | 0.920 | 0.990 | 0.707 | 0.993 | 0.710 | 0.876 | 0.853 | 0.956 | 0.649 | 0.929 |
| d2l14 | 0.910 | 0.980 | 0.936 | 0.955 | 0.860 | 1 | 0.701 | 0.993 | 0.734 | 0.822 | 0.87 | 0.974 | 0.642 | 0.92 |
| d2s14 | 0.910 | 0.980 | 0.933 | 0.964 | 0.830 | 0.990 | 0.585 | 1 | 0.748 | 0.822 | 0.801 | 0.956 | 0.705 | 0.916 |
| SD | 0.916 | 0.990 | 0.981 | 0.903 | 0.840 | 0.970 | 0.743 | 0.987 | 0.608 | 0.865 | 0.922 | 0.978 | 0.564 | 0.91 |
| SAM | 0.806 | 0.930 | 0.928 | 0.909 | 0.240 | 0.870 | 0.585 | 0.914 | 0.618 | 0.847 | 0.715 | 0.969 | 0.512 | 0.727 |

Table 4: Mutiple correlation coefficient between positive and negative accuracy for predicted dice across different models. The results indicate that as accuracy of prompts improve, the SAM based segmentation task performance improves as well.

| Segmentation tasks only | Kidney | TS: Liver | Shoulder | TS: Urinary Bladder | TS: Spleen | U/S GB | U/S Kidney |
|---|---|---|---|---|---|---|---|
| Multiple correlation coefficient | 0.92 | 0.67 | 0.99 | 0.95 | 0.67 | 0.95 | 0.98 |

### 4.2 Evaluation Metrics

Evaluation was done using different criteria for different tasks. For segmentation we assess the quality of correspondence as well as final segmentation using SAM. For computing prompt accuracy i.e. the accuracy of positive and negative prompts derived corresponding to that of template we use 2. For eg. If all positive prompts are inside the region of interest, then accuracy is 100% for positive prompts and vice-versa for negative prompts. Any positive prompts outside of ROI is considered as mislabeled and quantified as reduction in accuracy as follows:

$$Acc^+ = \frac{1}{n} \sum_{i=1}^{n} \mathbb{1}(Corr(T_i^+, I) \in I_M) \tag{2}$$

Where, $I_M$ is the ground truth segmentation mask. The accuracy for negative points can be computed similarly. Dice overlap metric is used for estimating the quality of segmentation using the derived prompts. For landmark correspondence task, normalized Euclidean distance (*dist*) between groundtruth and predicted points is reported as given by equation 3.

$$Loc_{error}(I) = \frac{1}{W} \sum_{i=1}^{n} dist(Corr(T_i^+, I), I_i^+)$$

(3)

### 4.3 Localization and segmentation results

Table 2, 3 and 5 along with Figures 4 and 2 summarize the metrics and performance for landmark correspondence and organ segmentation with various models. Table 4 demonstrates that accurate determination of positive and negative prompts is correlated to improvement in segmentation performance from SAM across all models. We notice that in all cases, DINO v2 and stable diffusion-based models perform best for these tasks with dice ranges of 0.62-0.90 and 0.57-0.88 respectively. DINO V2 almost outperforms DINO V1 in all tasks which is mostly likely due to well curated data and additional losses introduced in DINO V2 model. Moreover, from Table 2 and 5, we notice that there is no single model which outperforms all other models across different tasks, rather we have to choose the model optimal for the task at hand. In addition, a larger model does not necessarily provide the best performance consistently (e.g. d2s14 outperforms d2g14 for spleen segmentation). Additionally, we find that the models trained using self-supervised learning (DINO v2) perform better than their supervised counterparts (SAM, CLIP) and the performance is at par with SD models. Furthermore, we observe that for all segmentation tasks, except for Ultrasound kidney, the FM features combined with SAM, using only a single template image, outperformed UniverSeg [2] by a significant margin for both one shot and five shot segmentation - where the Universeg one-shot performance is significantly worse for all tasks.

This study has some shortcomings. One, we have not compared the performance of such single shot approach with supervised, in-domain task-specific models. Second, our approach is currently based on single slice, and it would be interesting to see how does this scale to 3D volume segmentation. The current study is limited to only few organs interspersed over various modalities and the next step would be to do an exhaustive evaluation over wider variety of cases expected in clinical practice (pathology, implants etc) in each of these modalities.

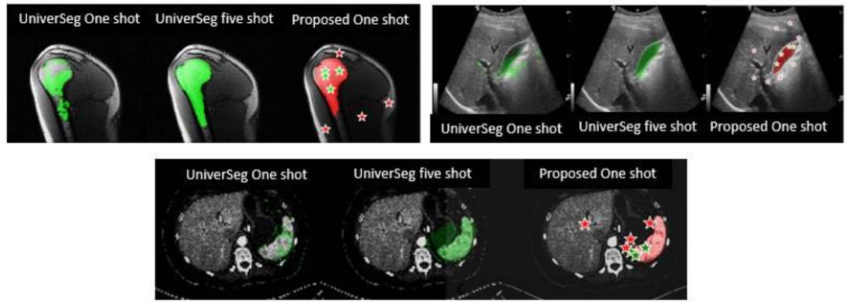

Figure 3: Segmentation performance comparison of the UniverSeg method with proposed methods.

Table 5: Dice metric reported for various segmentation tasks with various models. Acceptable dice varies based on the end-application based on organ segmentation.

| Dice Accuracy | Kidney (MR) | TS: Liver | Shoulder | TS: Urinary Bladder | TS: Spleen | U/S GB | U/S Kidney |
|---|---|---|---|---|---|---|---|
| d1b16 | 0.817 | 0.865 | 0.690 | 0.689 | 0.630 | 0.628 | 0.653 |
| d1b8 | 0.828 | 0.867 | 0.677 | 0.717 | 0.702 | 0.723 | 0.596 |
| d1s16 | 0.796 | 0.871 | 0.705 | 0.692 | 0.634 | 0.578 | 0.646 |
| d1s8 | 0.814 | 0.864 | 0.682 | 0.724 | 0.675 | 0.731 | 0.605 |
| d2b14 | 0.904 | 0.882 | 0.641 | 0.670 | 0.692 | 0.721 | 0.653 |
| d2g14 | 0.891 | 0.885 | 0.744 | 0.689 | 0.681 | 0.738 | 0.624 |
| d2l14 | 0.887 | 0.879 | 0.697 | 0.689 | 0.705 | 0.684 | 0.621 |
| d2s14 | 0.883 | 0.872 | 0.671 | 0.666 | 0.740 | 0.693 | 0.657 |
| SD | 0.878 | 0.872 | 0.649 | 0.726 | 0.645 | 0.741 | 0.546 |
| SAM | 0.667 | 0.821 | 0.186 | 0.597 | 0.604 | 0.563 | 0.502 |
| UniverSeg (1 shot) | 0.465 | 0.621 | 0.321 | 0.184 | 0.150 | 0.364 | 0.395 |
| UniverSeg (5 shot) | 0.804 | 0.727 | 0.718 | 0.476 | 0.508 | 0.675 | 0.687 |

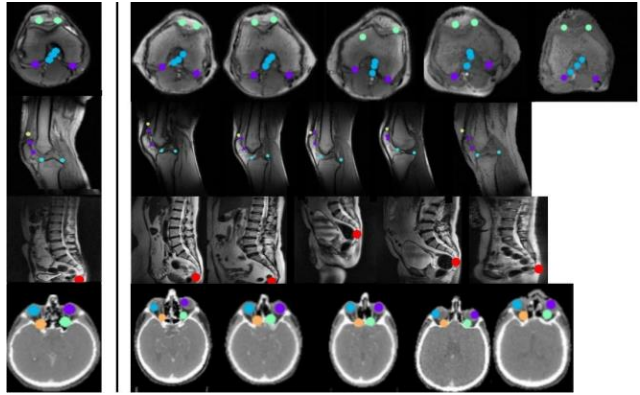

Figure 4: The landmark localization with best performing models for each task is shown here. The image on left of arrow is the template image with requested landmark. The images to the right are the different target images across subjects with different orientations, size and anatomy coverage. Notice reasonably good localization for various anatomical landmarks with a single foundation model based single shot learning.

## 5   CONCLUSION

In this study, we have examined whether models such as Vision Transformers and Stable Diffusion developed by the Computer Vision community extend to medical images. Over a variety of medical imaging modalities, anatomical regions and tasks without the benefit of in-domain training, we observe that such models can be indeed utilized for localization and segmentation tasks, provided appropriate feature correlation metric can be designed. We demonstrated the efficacy of the proposed chaining of correspondence with SAM, using a single exemplar, on twelve datasets spanning both localization and segmentation tasks and established its superiority over other recent methods [2] in the few shot realm.

## 6 Supplementary Material

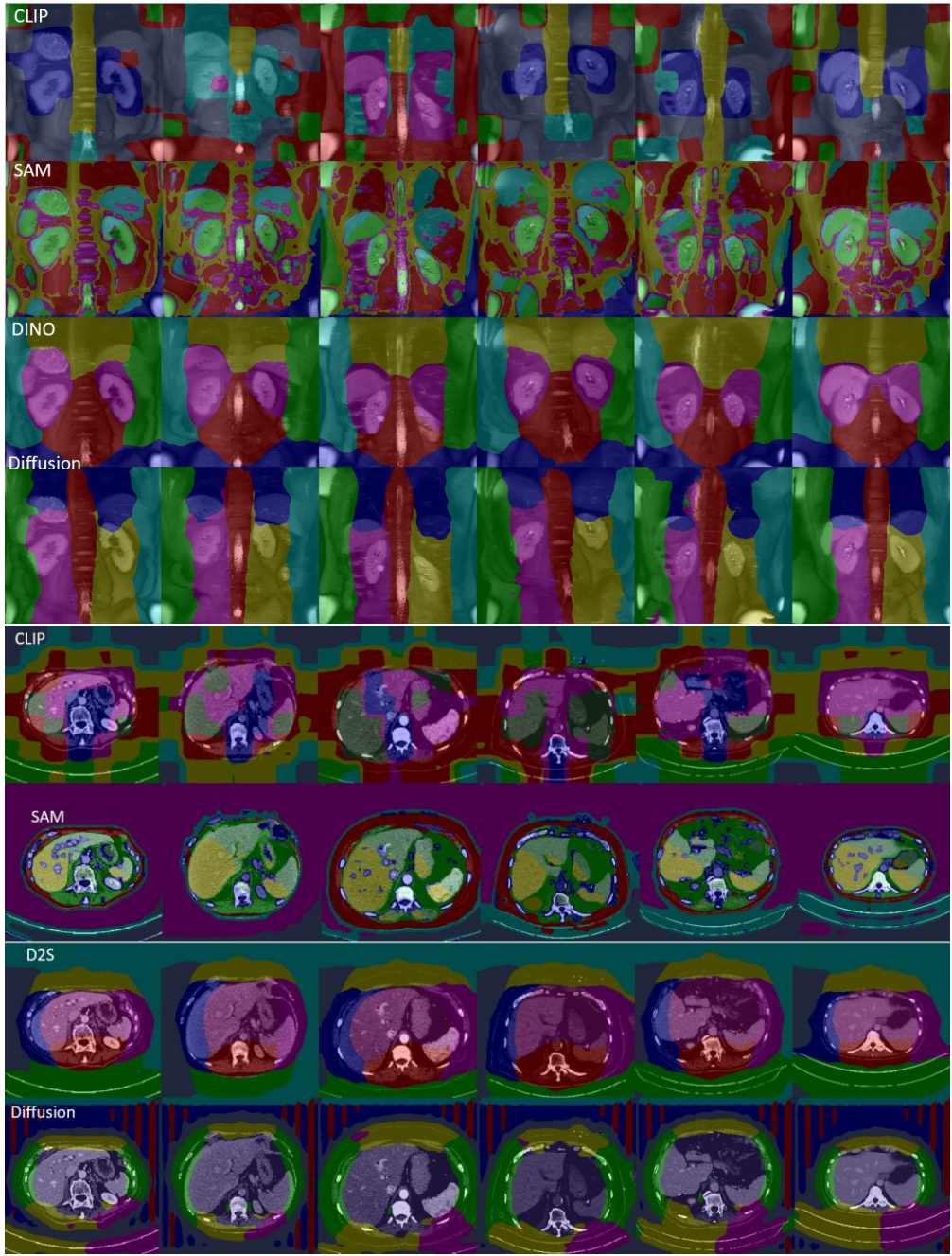

Figure 5: Results of co-lustering of pixel features from different models to understand efficacy of pixel feature based methodology for localization of region of interest. CLIP features are blotchy and don't always capture the semantic similarities (Notice kidney region has different semantic labels) . SAM – captures semantics well but gives many scattered similar regions which may not be similar from ROI standpoint. DINO (d2s14) – was able to group similar regions well – e.g. all kidneys same color – may not be at required level of granularity-e.g. other anatomies also colored pink in kidney example. Stable Diffusion (SD) – able to group regions but also shows specificity to symmetricity of image (note that kidney region left and kidney region right have different labels). Similarly in CT liver case, notice symmetricity in labels.

## 6.1 Dataset Description

Open-Source Datasets:

- CT Total Segmentator Dataset [13]: 1204 CT examinations and segmentations of 104 anatomical structures (27 organs, 59 bones, 10 muscles, 8 vessels). In this paper, we have used data from three organs: urinary bladder (40 cases), spleen (69 cases) and liver (74) for experiments.
- HaN Dataset [10]: Anonymized head and neck (HaN) images of 42 patients that underwent both CT and T1-weighted MR imaging. Segmentations of 30 organs at risk for the CT images.
- MRI kidney dataset [4]: 50 patients with chronic kidney diseases. T2 weighted abdominal MRI scans with kidney segmentation provided as ground-truth.

In-house datasets:

- MRI Shoulder dataset: MR SSFSE-localizer sagittal images from 25 subjects were used for experiments.
- MRI Knee dataset: MR localizers from two vendors (238 from Vendor1 and 45 cases from Vendor2) for knee landmark detection in axial and sagittal orientation. For sagittal image, three landmarks are : a. Meniscus points b. Patella point and c. Patellar tendon insertion point. For axial images, the three landmarks are: a. Patella surface b. posterior femoral condyle and c. inner surface of the lateral femoral condyle.
- MRI Spine dataset: Sagittal T1 weighted images from 182 subjects collected from two different clinical sites are used for spine vertebrae localization in lumbar and cervical stations. One template image is chosen for both the stations and centroid of each vertebra is marked to generate pixel level features.
- Ultrasound abdominal data: Appropriate Ultrasound cross-sections of routine evaluations of the abdomen using a single vendor scanner from multiple clinical sites under an IRB approved data sharing agreement, for gall-bladder and kidney segmentation, was used.

## 6.2 Additional Experiments and Results

In this section we present some more results from our experiments to understand and visualize the efficacy of different model in deriving meaningful dense features, to explore the different ways in which these dense features can be utilized for identifying RoI as well as the robustness of the various models to changes in image poses.

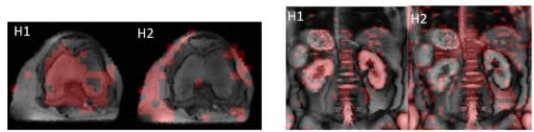

Figure 6: (a) Attention maps from DINO for MR knee axial image (b) DINO attention maps for MR kidney images

### 6.2.1 Attention based localization using DINO V1

Using attention maps from various attention heads from DINOv1 model (specifically the d1b8) to identify regions of interest showed some interesting results for natural images [3]. Similar attention head output was evaluated in MRI Knee Axial and MRI Kidney images to ascertain if they provide good anatomical localization. Fig 6 shows the attention map from DINO v1(d1b8) model for heads H1 and H2. We first note that the semantic region coverage of attention maps for medical images is not as clean as for natural images (See [3], fig 4) . We notice that the first attention map (H1) for both medical images in provides a rough segmentation of the knee femoral bone and the kidneys respectively whereas the second attention map (H2) captures the region outside the organ of interest. It might be possible to further heuristically cleanup these maps to obtain a rough region localization maps. However, such a map will not provide us control on the region on interest (e.g fetch the posterior end of femoral condyle) and hence we discarded this approach from further consideration.

6.2.2   Similarity based mask generation

We also explored similarity based mask generation as an alternative to the proposed method of chaining the derivation of corresponding prompts using dense features with SAM for segmentation tasks. In this approach we mark all the pixels in the target image whose similarity with any of the template image positive prompt pixels is high i.e. pixels whose similarity is in the 80th percentile among all target image pixels are considered. Samples of the results of such masks generated from d2g14 are shown in fig 7. The choice of the model was based on visual examination of results where d2g14 results look better for most cases. We observe that for images where the RoI are well delineated the similarity mask is able to cover the RoI reasonably well and may be refined using mask refinement algorithms. However, the method is heuristic in nature i.e. the setting of parameters e.g. the similarity threshold limit needs careful consideration, and we are actively pursuing on automatically determining this per task.

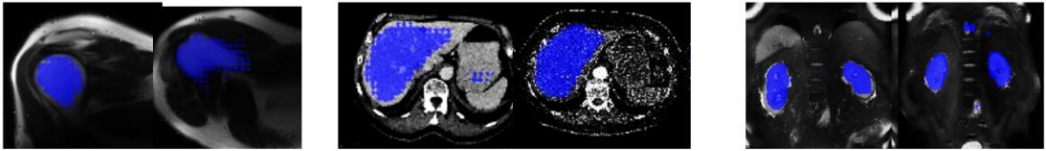

Figure 7: Similarity based masks using d2g for (a) MR Shoulder (b) CT Liver (c) MR Kidney

6.2.3   Clustering of pixel-level feature embeddings to verify semantic segmentation of similar regions

We co-clustered the features (using k-means) on multiple target images and overlaid them on target images. The quality of clustering was ascertained by visualizing the semantic regions to have similar labels (i.e. colors) across the target images. This was evaluated on Total Segmentator Liver and MR Kidney datasets.

Figure 5 shows the semantic segmentation obtained by clustering target image features for various models. Results indicate that this approach is feasible; especially with DINO, SAM, stable diffusionbased and to some extent by CLIP models. However, the quality of clustering using CLIP seems inferior to the other models and thus was not considered for further analysis.

6.2.4   Robustness of DINO and SD to image pose variations

We also evaluated the robustness of the two models (d2s14 and SD) to understand the impact of local context in their performance. Accordingly, as an extreme example, the template was flipped along left right or up-down and corresponding landmark provided. As seen in Figure 8, DINO v2(d2s14) model was found to be more reliable to localize the landmark despite image orientation changes. This is critical since in medical imaging image poses can be variable across scans and as such DINO v2 will be more appropriate.

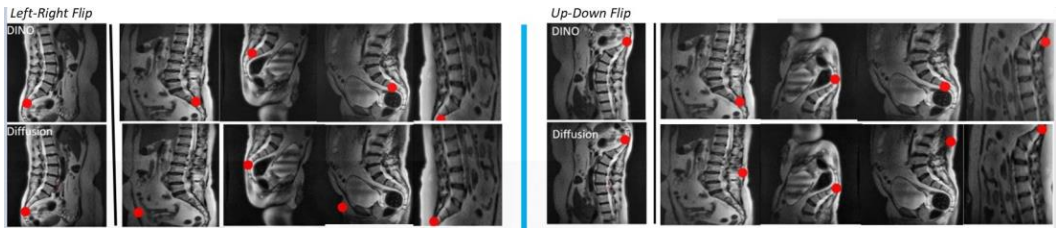

Figure 8: The template image (left most image) was flipped, while some of target images were kept intact or flipped to test robustness of two models. DINO (top row) was found to be more consistent in localization of landmark, compared to stable diffusion (bottom row).

[14] Junyi Zhang, Charles Herrmann, Junhwa Hur, Luisa Polania Cabrera, Varun Jampani, Deqing Sun, and Ming-Hsuan Yang. A tale of two features: Stable diffusion complements dino for zero-shot semantic correspondence. *arXiv preprint arXiv:2305.15347*, 2023.12